\documentclass[10pt,twocolumn]{article} 
\usepackage{simpleConference}
\usepackage{times}
\usepackage{graphicx}
\usepackage{amssymb}
\usepackage{url,hyperref}
\usepackage[center, FIGTOPCAP]{subfigure}
\usepackage{cite}
\usepackage{multirow}
\usepackage{siunitx}

\begin{document}

\title{A general learning system based on neuron bursting and tonic firing}

\author{Tim Lui \\
\\
\today
\\
\\
lui.thw@cantab.net  \\
}

\maketitle
\thispagestyle{empty}

\begin{abstract}\label{sec-abstract}
This paper proposes a theoretical framework for the biological learning mechanism as a general learning system. The proposal is as follows. The bursting and tonic modes of firing patterns found in many neuron types in the brain correspond to two separate modes of information processing, with one mode resulting in awareness, and another mode being subliminal. In such a coding scheme, a neuron in bursting state codes for the highest level of perceptual abstraction representing a pattern of sensory stimuli, or volitional abstraction representing a pattern of muscle contraction sequences. Within the 50-250 ms minimum integration time of experience, the bursting neurons form synchrony ensembles to allow for binding of related percepts. The degree which different bursting neurons can be merged into the same synchrony ensemble depends on the underlying cortical connections that represent the degree of perceptual similarity. These synchrony ensembles compete for selective attention to remain active. The dominant synchrony ensemble triggers episodic memory recall in the hippocampus, while forming new episodic memory with current sensory stimuli, resulting in a stream of thoughts. Neuromodulation modulates both top-down selection of synchrony ensembles, and memory formation. Episodic memory stored in the hippocampus is transferred to semantic and procedural memory in the cortex during rapid eye movement sleep, by updating cortical neuron synaptic weights with spike timing dependent plasticity. With the update of synaptic weights, new neurons become bursting while previous bursting neurons become tonic, allowing bursting neurons to move up to a higher level of perceptual abstraction. Finally, the proposed learning mechanism is compared with the back-propagation algorithm used in deep neural networks, and a proposal of how the credit assignment problem can be addressed by the current proposal is presented.

\end{abstract}

\section{Introduction}\label{sec-introduction}

Over the past decade, there were majour advances in the field of artificial intelligence and machine learning. Biologically inspired deep neural networks have gained increasing popularity with widespread use~\cite{Lecun2015a}. The back-propagation algorithm~\cite{Rumelhart1986, LeCun1998} and rectified linear units~\cite{Nair2010} allow deep neural networks with many layers of non-linearity to be trained successfully, and the explosion of data generated by the Internet allows these deep neural networks to be trained without overfitting. Recent advances in deep reinforcement learning allow intelligent agents to learn by interacting with the environment without supervised training~\cite{Mnih2015a, Mnih2016}. Problems previously unsolvable by rule-based systems with brute force search can now be tackled with deep reinforcement learning, such as mastering the game of Go~\cite{Silver2016, Silver2017}.

Despite the recent advances, the state of the art learning systems are still considered as narrow learning systems, with limited ability to generalise their learnings across multiple domains. An agent trained to play a particular Atari game has limited ability to transfer its learning to play another Atari game~\cite{Rusu2016}. AlphaGo is able to surpass the human level of performance in the game of Go after many games of self-play~\cite{Silver2017}, but it is unable to communicate with people similarly to Apple Siri. A general learning system with a single architecture and learning rule capable of mastering different skill sets across multiple domains is still an active area of research~\cite{Goertzel2014}.

A general learning system should be able to focus computational resources to search for optimal solutions in novel circumstances, while keeping the found solutions for familiar circumstances in memory. Humans, as well as many animal species, seemingly have two distinctive modes of information processing, with one mode resulting in awareness, and the other being automatic and subliminal, as if they are cortical reflexes~\cite{Koch2001}. It is known that people can drive when in deep sleep, an extreme variant of sleepwalking~\cite{Southworth2008}. If the brain can perform such complex sensorimotor tasks entirely subliminally, and reacting appropriately to complex visual scenes without awareness, then what is the function of awareness? It seems that the key difference being that the learning of new tasks often requires awareness before automatic subliminal processes can take over, and people can not reason or think when they are sleepwalking. Such evolutionary design has the advantage of allowing the brain to attend to the most novel aspect of sensory stimuli, while the familiar stimuli are processed subliminally. In this paper, it is proposed that there is a distinctive neural state to represent novelty for the information processing system of the brain that results in awareness.

\section{Novelty representation}\label{sec-novelty-representation}

There is a difference between novelty and sensory saliency. All neurons respond to sensory saliency, however sensory saliency alone is a necessary but insufficient condition for novelty. When the visual stimuli of an object fall onto the retina, they trigger the activity of neurons in the visual cortex that code for different aspects of visual saliency. In the current understanding of the information processing hierarchy of the visual cortex, neurons in the lower visual cortical areas send action potentials, or spikes, to the higher areas, allowing lower level features to be integrated into higher level features, from simple edges at the primary visual cortex (V1), to shapes at V2, and motions at V5, etc ~\cite{Size2009}. Eventually neurons at the inferior temporal gyrus code for an entire object, such as faces. However, the act of seeing an object does not necessary result in awareness. Under the state of highway hypnosis, a driver may see many vehicles passing by without perceiving anything of them with awareness~\cite{Williams1963}. 

If a combination of salient sensory stimuli represents a familiar object or event, it is not necessary to attend to those stimuli in great detail. Novelty thus requires a combination of salient stimuli that does not have a prior neural representation. Under the state of highway hypnosis, the driver may not be able to perceive the regular familiar vehicles passing by, but most likely will if there is a pink fire truck. Therefore, there must be two distinctive information processing methods for novel and non-novel stimuli in the brain, such that they result in awareness and subliminal responses respectively.

It is proposed that such distinctive neural state for the processing of novel stimuli that results in awareness is in fact the bursting mode of many neuron types found in the brain, such as pyramidal neurons in the cortex~\cite{Baranyi1993, Gray1996, Gray1997}, pyramidal neurons in the hippocampus~\cite{Traub1981, Miles1986}, midbrain dopaminergic neurons~\cite{Wang1981, Hyland2002}, and thalamocortical relay neurons in the thalamus~\cite{Jahnsen1984, Williams1997}.

\subsection{Neuron bursting mode}

\begin{figure}[htbp]
 	\centering
 	\subfigure[Novel]{
	\includegraphics[width=0.9\linewidth]{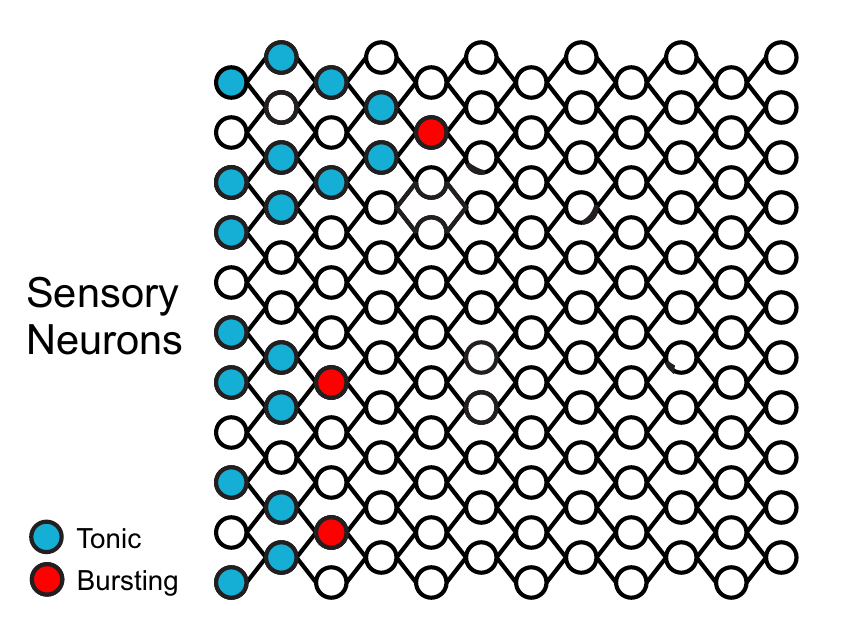}
	}
 	\hfill
	\subfigure[Familiar]{
	\includegraphics[width=0.9\linewidth]{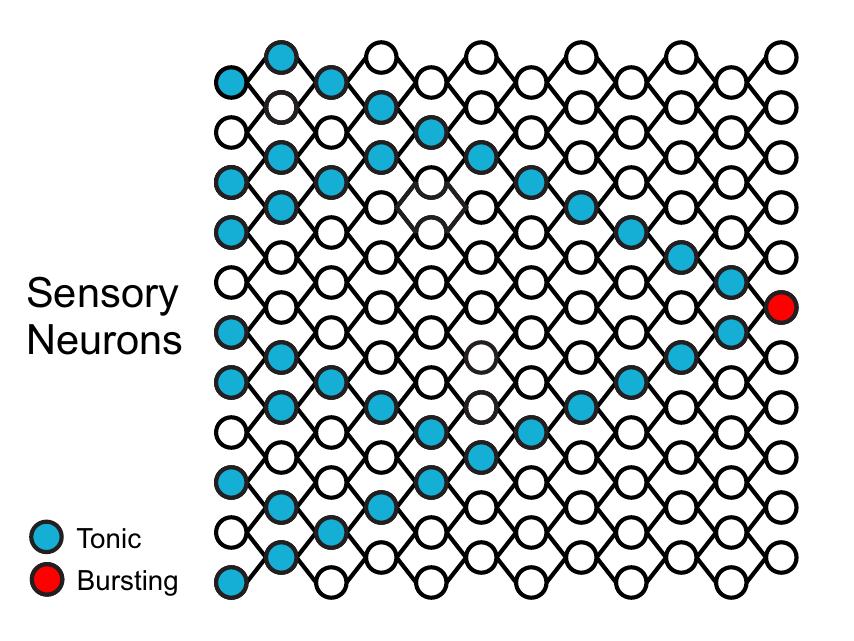}
	}
\caption{Neuron coding for a) novel and b) familiar sensory stimuli, showing more neurons in bursting mode for novel stimuli}\label{fig:nolvety-representation}
\end{figure}

For the neuron types described above, they have two distinctive modes of firing pattern, tonic mode which sends regular single spikes to downstream neurons, and bursting mode which sends a burst of spikes. It was found that neurons typically switch between these two modes of firing, but under what conditions for such switch to happen remain unclear~\cite{Cooper2002, Hyland2002}. It is proposed that only neurons coding for the highest level of perceptual abstraction for the current sensory stimuli would enter into bursting mode, and neurons in this state would trigger a series of events in the brain that results in awareness. Under this hypothesis, the visual stimuli of a vehicle would result in a cascade of neuron activity in the visual cortex, with most of these neurons being in tonic mode, but a few would be active in bursting mode. The majourity of neurons in tonic mode code for lower level visual features, while those in bursting mode code for the highest level of abstraction. The bursting neurons may code for different high level features of the current stimuli, such as the vehicle type, the brand, or the direction of travel. Thus, novelty is represented by the novel combination of high level features coded by the bursting neurons. Novelty is not a binary property but as a continuous scale. The level of novelty is coded by the number of neurons in bursting mode. If the object has a high level of novelty, a larger number of neurons enter into bursting mode (figure~\ref{fig:nolvety-representation}), as the cortex has yet developed the appropriate higher level perceptual abstraction to represent the entire observation.

\begin{figure}[htbp]
 	\centering
 	\subfigure[Sensory percept]{
	\includegraphics[width=0.9\linewidth]{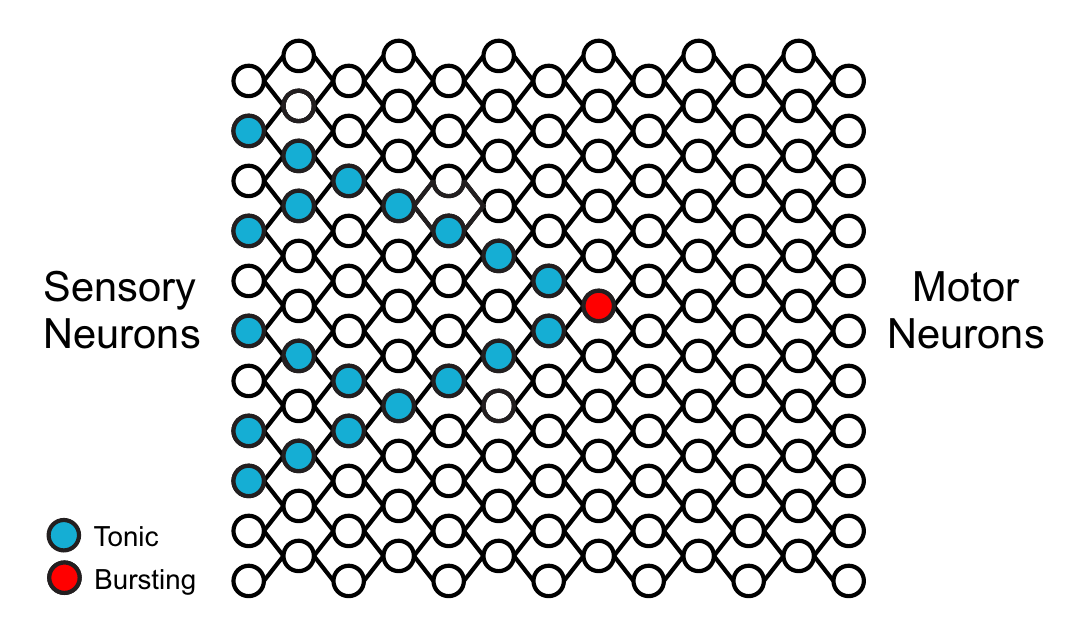}
	}
 	\hfill	
	\subfigure[Motor volition]{
	\includegraphics[width=0.9\linewidth]{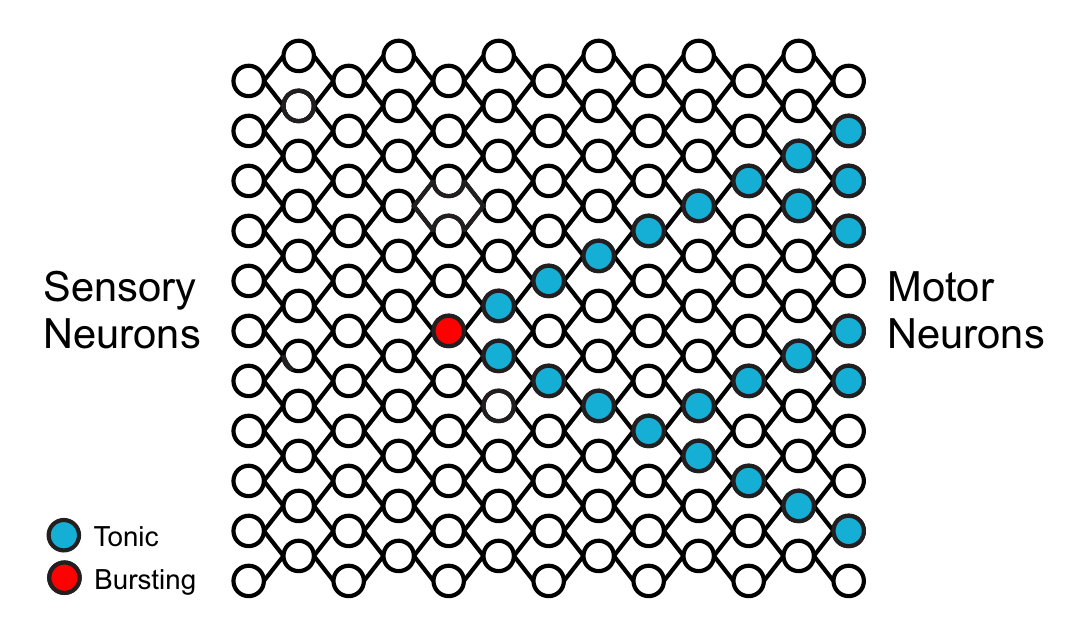}
	}
 	\hfill	
	\subfigure[sensorimotor control with awareness]{
	\includegraphics[width=0.9\linewidth]{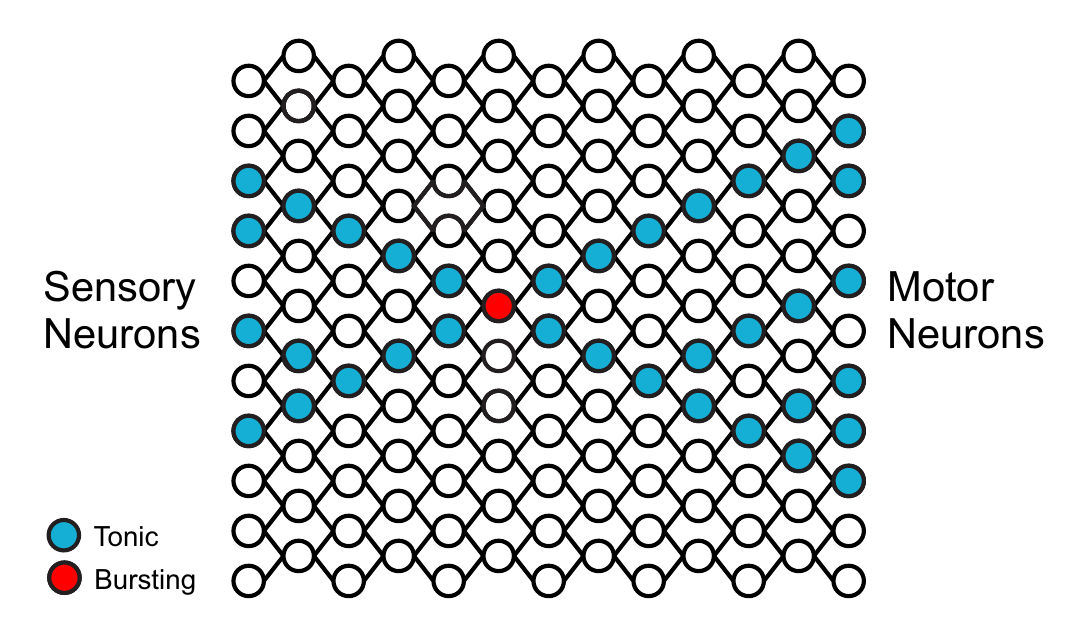}
	}	
 	\hfill	
	\subfigure[Subliminal sensorimotor control]{
	\includegraphics[width=0.9\linewidth]{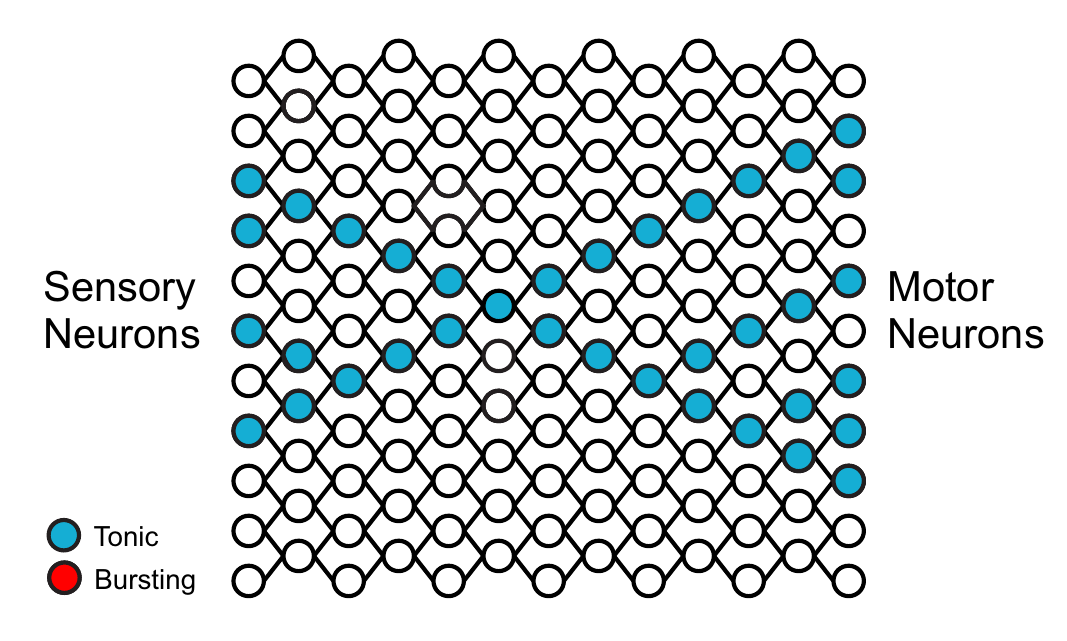}
	}	
\caption{Neuron coding for different high level abstractions}\label{fig:motor-volition}
\end{figure}

On the motor side, bursting neurons represent the highest level of abstraction for motor commands. As illustrated in figure~\ref{fig:motor-volition}, on the sensory side, a series of neurons in tonic mode lead to the bursting activity of the neuron coding for features of the highest level of sensory percept; on the motor side, bursting neurons coding for the highest level of motor volition lead to the tonic activity of a series of neurons coding for a particular muscle contraction sequence.

Under this hypothesis, all bursting neurons result in awareness, which may be coding for sensory percept, motor volition, or both. Sensory stimuli could lead to bursting activity of neurons coding for the highest level of sensory percept, and this in turn may have immediate downstream tonic activity in motor neurons (figure~\ref{fig:motor-volition} part c). For highly familiar circumstances, the activity from sensory neurons to motor neurons could be entirely tonic and therefore subliminal. In fact, evidence suggests that all motor acts require a degree of continuous sensory feedback to ensure smooth and successful execution, and such process can occur subliminally~\cite{Diedrichsen2010}. 

Such coding scheme allows the brain to attend to the most novel and abstract representation of the current sensory stimuli, thus allowing more efficient use of computational resources. For familiar objects or events, not only has the brain already developed the appropriate perceptual abstractions, it may have also developed the corresponding motor commands given the sensory stimuli. The driver under the sate of highway hypnosis is constantly adjusting the steering wheel in response to the visual stimuli of the road, without perceiving the constantly varying curvature of the road. This is not always the case, when first learning to drive, the curvature of the road and the act of steering the steering wheel are both perceived by the driver with high awareness. However, over time as one becomes familiar with driving, they become more subliminal. In other words, not only do familiar sensory stimuli become more subliminal, it is also true for familiar motor commands as one acquires the motor skills. The majourity of neurons previously in bursting mode are switching to tonic mode as one learns. Neurons in tonic mode can carry their activity all the way from sensory neurons to motor neurons, resulting in subliminal sensorimotor control. The mechanism of learning and neural plasticity is further discussed in section \ref{sec-learning}.

\subsection{Bursting inhibition mechanism}

Questions remain on the exact mechanism which neuron bursting mode is modulated to represent novel precepts. The novelty representation hypothesis states that a neuron would take the tonic state if its activity can excite some downstream neurons to burst, which in turn depends on the activity of other neighbouring neurons from which the downstream neuron receives its dendrite inputs. Cortical pyramidal neurons receive inputs from the basal and the apical dendrites, and inputs from the apical dendrite is what modulate its bursting activity~\cite{Larkum1999}. A large proportion of cortical neurons are inhibitory interneurons, accounting for around 20\% of all cortical neurons~\cite{Meinecke1987}. They can inhibit more than 50\% of pyramidal cells within \SI{100}{\micro\metre} and receive excitatory inputs from a large fraction of them~\cite{Fino2011}. In many cases interneurons were found to be connected to every local pyramidal neuron sampled~\cite{Packer2011}, as well as having reciprocal connections to each other~\cite{Galarreta2002}. 

\begin{figure}[htbp]
 	\centering
 	\subfigure[]{
	\includegraphics[width=0.65\linewidth]{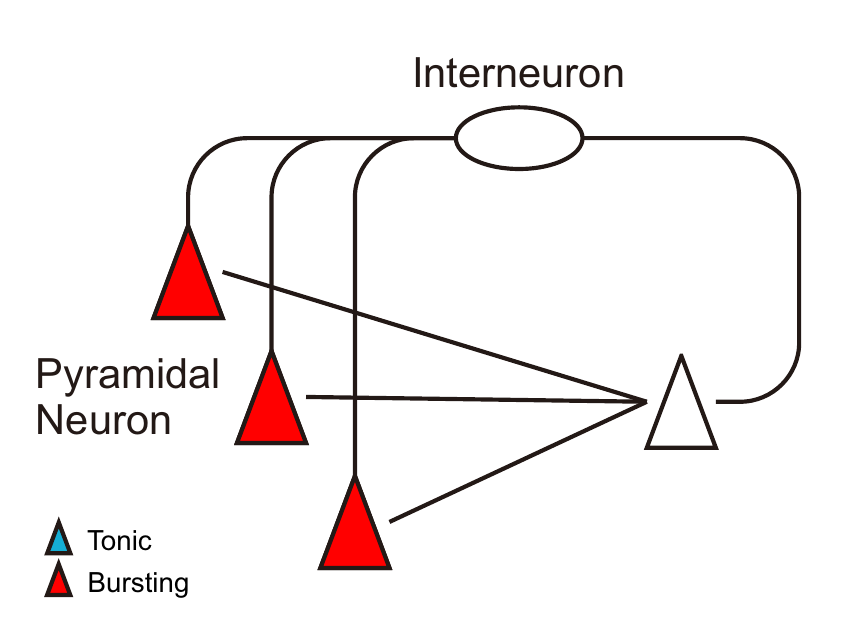}
	}
	\subfigure[]{
	\includegraphics[width=0.65\linewidth]{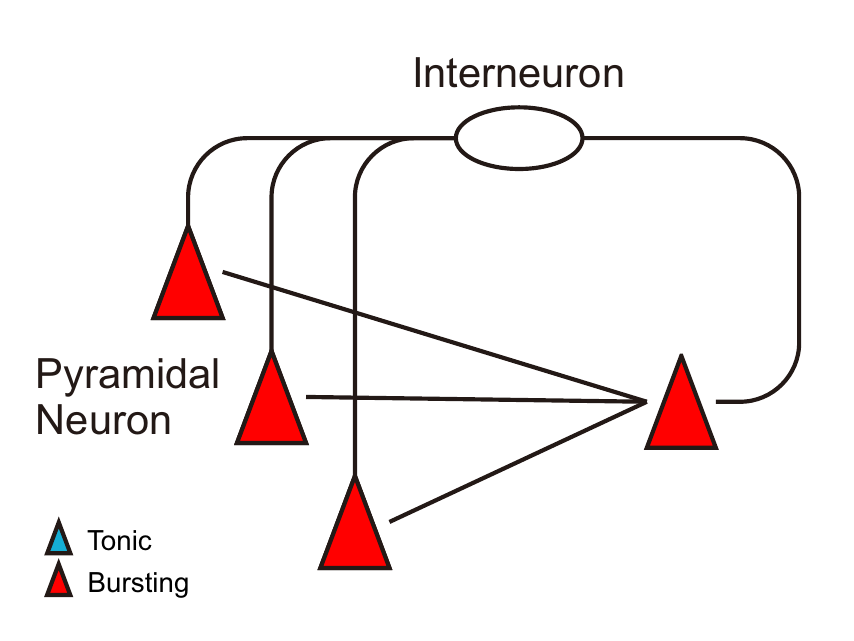}
	}
	\subfigure[]{
	\includegraphics[width=0.65\linewidth]{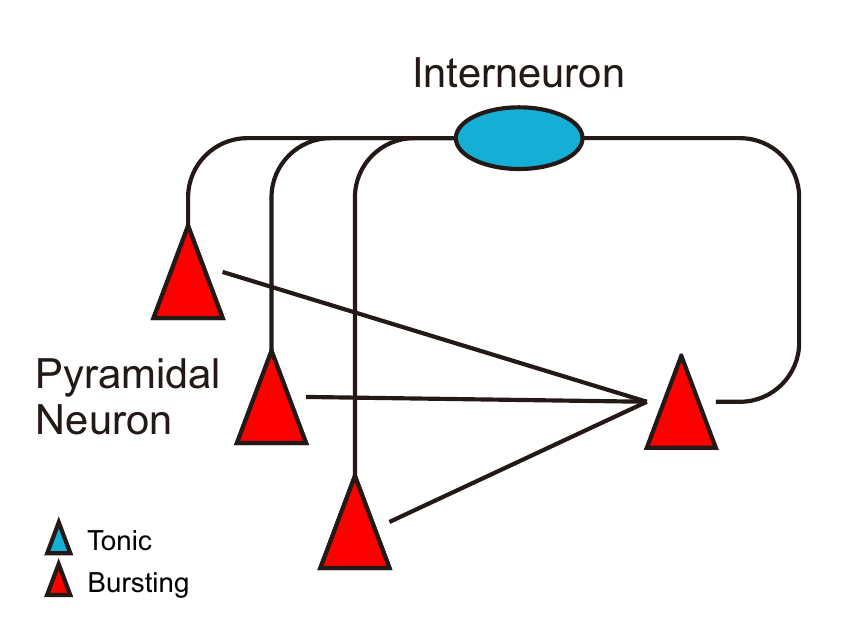}
	}
	\subfigure[]{
	\includegraphics[width=0.65\linewidth]{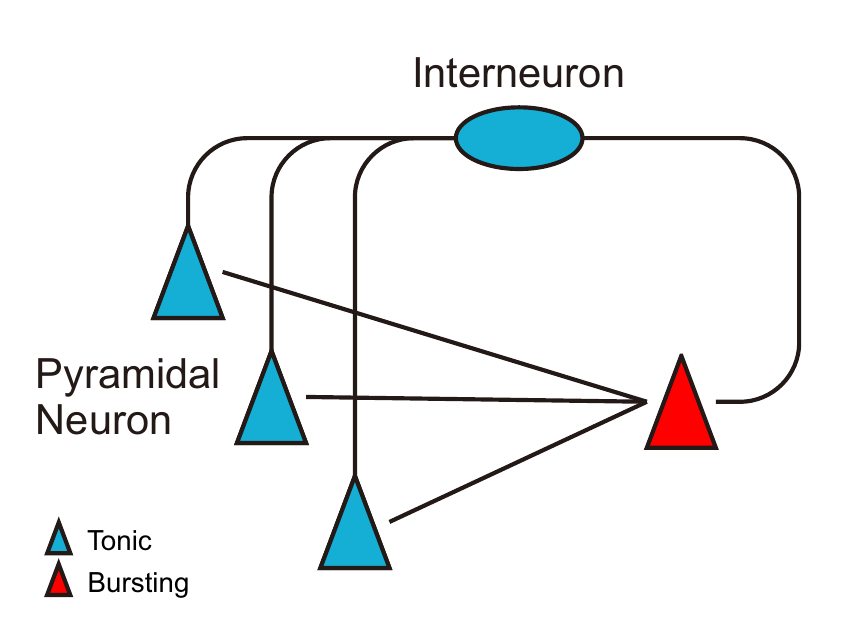}	
	}	
\caption{The bursting inhibition mechanism. a) Upstream pyramidal neurons enter bursting mode, b) causing the downstream pyramidal neuron to burst. c) Downstream inhibitory interneuron becomes active. d) Inhibitory feedback connections to apical dendrite of upstream pyramidal neurons cause them to enter tonic mode.}\label{fig:bursting-inhibition}
\end{figure}

It is proposed that neurons always enter bursting mode when being excited by upstream neurons, unless receiving inhibitory apical inputs from downstream inhibitory interneurons. Figure~\ref{fig:bursting-inhibition} illustrates the process of the bursting inhibition mechanism.

The activity of a neuron depends on the synaptic connection strength between itself and its upstream neurons. The connection strength between an upstream and a downstream neuron is only increased if the combination of upstream activities has been repeatedly observed in prior experiences, as discussed further in section~\ref{sec-learning}. This means that the downstream neuron's activity is a higher level abstraction of the combination of activities of upstream neurons. From an information processing prospective, if a higher level abstraction of current activity is found, there is no longer a need to process the lower level information, and the activity of upstream neuron becomes tonic. In other words, a neuron always switches from bursting to tonic state if its own activity can be explained by the activity of a downstream neuron. Recursively, all the lower level neurons switch to tonic state, and only the neuron coding for the highest level of abstraction remains bursting, forming a tree-like structure, with only the top neuron of the tree bursting.

There are numerous pieces of neurophysiological evidence supporting the current hypothesis. Local inhibition is often observed to be lagging local excitation with a delay of a few milliseconds. This was found in the auditory cortex of rats~\cite{Wehr2003, Wu2008}, somatosensory cortex after whisker deflections~\cite{Swadlow2002, Wilent2005}, as well as the visual cortex after a light flash stimulation~\cite{Liu2010}. Interneuron inhibition was found to play a critical role in the generation of fast gamma oscillations in the cortex and the hippocampus, by pacing and synchronising the activity of large population of neurons ~\cite{Cobb1995, Cardin2009, Sohal2009}.

The illustration of figure~\ref{fig:bursting-inhibition} models the mechanism of bursting inhibition on the interaction of individual excitatory pyramidal neurons and inhibitory interneurons. However, it should be noted that the current hypothesis can be applied to any level of neuron circuitry which exhibits excitatory and reciprocal inhibitory control. There is reason to believe that the correct level of organisation of the proposed hypothesis is at the level of cortical columns. Specifically, it is the cortical column which codes for a percept, and it is the interaction between adjacent cortical columns as well as across the cortex that allows for bursting inhibition.

In the cortex, neurons are organised into 6 separate layers. Sensory inputs and thalami relays usually synapse at layer IV, while motor commands and other thalami outputs originates at layer VI~\cite{Felleman1991}. Cortical columns also send information to each other via ascending and descending pathways between higher and lower cortical areas. The thalamus has relay nuclei organised in parallel columns, with each nucleus connecting to a separate cortical column, as if the thalamus is the 7th layer of the cortex~\cite{Steriade1988}. In addition to a one-to-one connection to a cortical column, some thalamus nuclei also have one-to-many diffuse connections to large portions of the cortex~\cite{Jones2007}. The most interesting fact relevant to the current hypothesis is that cells at the reticular complex at the thalamus surface can be excited by bidirectional pathways from and to the cortex, and send inhibitory outputs to the origin of the pathway. This suggests that the bursting inhibition mechanism may be mediated by the thalamus nucleus to the corresponding cortical column.
\section{Percept integration}\label{sec-percept-integration}

\subsection{The binding problem}
Assuming different neurons code for different features of an object and from different sensory modalities, such as shape, size, sound, etc; the binding problem is how the brain determines what collection of features are related to each other, and what should be perceived separately~\cite{Roskies1999}. In a busy city street, auditory stimuli from nearby construction site and road traffic can both be processed by the brain simultaneously: what neural mechanism ensures that the two streams of sound are perceived separately, and what neural mechanism allows the sounds from different vehicle types to be perceived as a single percept of traffic sound?

One observation is that our ability to perceive subtle differences between objects depends on our familiarity with the objects, as well as the level of effort made to perceive the difference. The cross-race effect~\cite{Behrman2001} is a good example of such a phenomenon. It is the tendency for people to find it difficult to discriminate faces from other races. ``Why do all Chinese people look the same?''

The ability to determine perceptual relations and separations are intertwined. The ability to establish relations between features of the same object enhances the ability to separate such object from other similar objects. The set of facial features required to separate foreign faces is different from the set of facial features required to separate faces from one's own race. The ability to separate the sound from traffic and construction requires a prior perceptual map of traffic and construction sounds. The ability to separate the sound from different vehicle types requires one to be familiar with different vehicle sounds.

Temporal order plays an important role in percept integration. Eye fixation allows features of the same object to fall onto the retina simultaneously or in quick successions. Experimental results show that the minimum integration time of experience is within 50-250 ms, centred at 100 ms~\cite{Blumenthal1977}, beyond which events are treated separately. In addition to eye fixation as a selective attention mechanism, the stream of thoughts further described in section~\ref{sec-episodic-memory} is another major mechanism to regulate feature binding. The sequential activation of bursting neurons by the hippocampus allows a deeper degree of percept binding beyond the duration of eye fixation, and allows for selective attention for all sensory modalities.

However, concurrent neuron activation still cannot provide the complete solution to the binding problem. If all features within the minimum integration time of experience are bound, then how could the brain separate concurrent stimuli coming from separate sources?

\subsection{Binding by synchrony}\label{bursting-neuron-resonance}
What features can be bound together must also depend on the underlying perceptual relations in addition to simple concurrency. Such map of perceptual relations must be inherent to the individual, as what can be treated separately differs from person to person. In other words, perceptual relations are learnt over time, and are represented by the synaptic connection strength between cortical neurons. What features can be bound together in the same integration time window of experience also depend on the existing cortical connections.

Section~\ref{sec-novelty-representation} discussed that the bursting neurons represent the highest level of perceptual abstraction for a pattern of sensory stimuli. In addition, there could be more than one active bursting neuron with each coding for different high level features of the same physical object.

When there are two sets of sensory stimuli coming from two separate physical objects, the brain needs to separate bursting neurons linked to separate objects into separate ensembles. The temporal binding theory suggests that neurons coding different features of the same object are bound by firing in temporal synchrony with a precision in millisecond range, whereas no synchronisation should occur between neurons coding for different objects, resulting in co-activation of multiple ensembles with differing firing rates~\cite{Engel2001}.

\begin{figure}[htb]
 \centering
 \includegraphics[width=\linewidth]{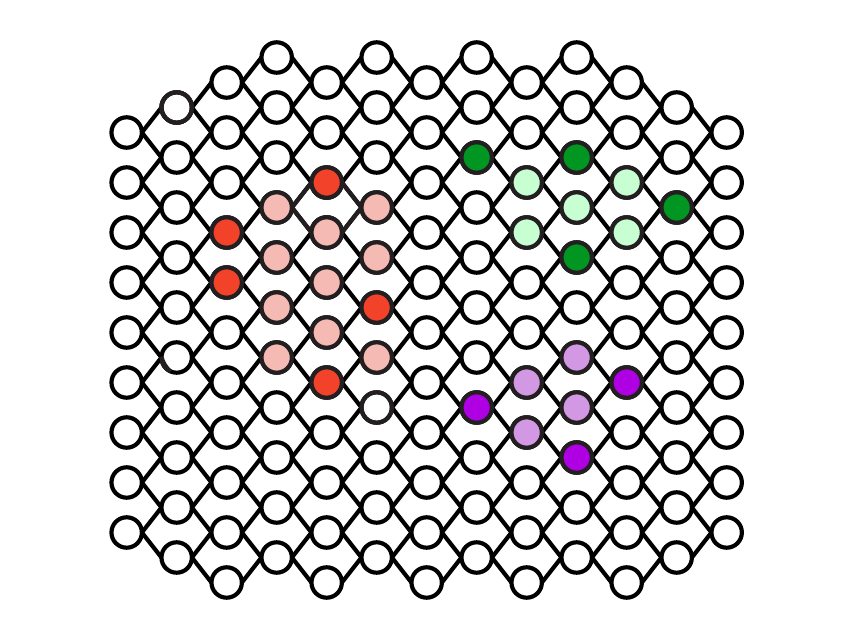}
 \caption{Neurons forming three competing synchrony ensembles, illustrated by neurons in tonic (lighter colour) and bursting (darker colour) mode. Spatial clustering is just for illustration purpose and ensembles may form across different regions in the cortex}\label{fig:synchrony-ensembles}
\end{figure}

For any two bursting neurons, if they have close underlying perceptual relations, there are more shared connections in active tonic mode coding for low level features common to both. In addition, the thalamus provides relay circuitry to allow neurons in different cortical regions to form synchrony ensembles and allow feature binding from different sensory modalities. If two bursting neurons code for features of two separate objects, there are fewer shared connections and thus a lower probability of synchronisation. Within the 50-250ms minimum integration time of experience, multiple neuron ensembles could form (figure~\ref{fig:synchrony-ensembles}).

Numerous pieces of evidence were found to support the temporal binding theory. Repeated experiments on cats' visual cortex found that spatially separated neurons respond in synchrony when activated by a single light bar, but form two separate ensembles when activated by two overlapping light bars with different orientations~\cite{Engel1991, Freiwald1995}. The firing frequency of these ensembles were found to be in 40-60Hz and no phase difference between neurons was found~\cite{Gray1989a}. Another experiment found that neuron synchrony occurs when responding to the contours of the same surface, but not when the contours belong to different surfaces~\cite{Castelo-Branco2000}. Similar experiments have been replicated on macaque monkeys~\cite{Williams1997, Kreiter1996}. In addition, coherence of local field potentials has also been found between sensory, motor, parietal, and other higher-order cortical areas~\cite{Bressler1993, Roelfsema1997}. In addition to stimulus driven synchrony, top-down expectation driven synchrony was also found during motor-preparation in monkeys. In a delay-reaching task, the level of network synchrony was found to be correlated with the growing stimulus expectancy, and this was able to predict the performance and the reaction time of the monkey~\cite{Riehle2000, Riehle1997}.

When there are multiple synchrony ensembles active concurrently, eventually one will become dominant and be actively maintained. This depends on the relative size of the ensembles, the firing rate of the ensembles, as well as neuromodulation discussed further in section \ref{sec-neuromodulation}. Such a hypothesis was also made under the temporal binding theory~\cite{Engel2001}.

Another related theory worthy of discussion is the adaptive resonance theory (ART)~\cite{Grossberg2013}. ART suggests that bottom-up neurons coding for a sensory pattern are matched against top-down neurons coding for a learnt expectation. A successful match between bottom-up and top-down neurons results in resonance, whereas a big enough mismatch, defined by the vigilance parameter, leads to an orientation response including a memory search and learning of new recognition categories. Similarly to ART, the current proposal also involves a matching process between current sensory stimuli and prior learnt patterns, as described by the bursting inhibition mechanism in figure~\ref{fig:bursting-inhibition}. However, in contrast to ART, the current proposal suggests that it is the mismatch that results in neuron burst firing and synchrony of bursting neurons, whereas the matched pattern is masked and handled subliminally.

\section{Episodic memory}\label{sec-episodic-memory}

One of the processes of the brain is the recall and formation of episodic memory. Episodic memory allows any combination of sensory percepts or motor volitions to be bound from all sensory modalities across the entire cortex. It allows experiences to be recorded in a film like manner~\cite{Tulving2002}. It is established that the hippocampus and the wider medial temporal lobe is the key brain structure for the coding of episodic memory~\cite{Greenberg2003}.

It is proposed that only neurons in bursting mode from the dominant neuron ensemble could trigger episodic memory recall and formation. This means that the hippocampus does not record all aspects of an experience, but only the most novel aspects. During episodic memory formation, the hippocampus records the sequence of activity of bursting neurons, and during memory recall, the hippocampus replays the same sequence by reactivating the corresponding neurons to burst. The hippocampal theta wave with a frequency of 4-7 Hz is suggested to be the clock cycle of episodic memory~\cite{Cantero2003}, and it was found that the high frequency gamma wave is phase locked to the theta wave in humans~\cite{Canolty2006a}. This finding matches with the 50-250 ms minimum integration time of experience~\cite{Blumenthal1977}. It was suggested that multiple memory patterns can be nested in each gamma (\(\sim\)40Hz) subcycle of the theta cycle. This also explained the Miller's law of 7\(\pm\)2 items of limited capacity of working memory~\cite{Miller1956}. 

\begin{figure}[htbp]
 	\centering
 	\subfigure[]{
	\includegraphics[width=\linewidth]{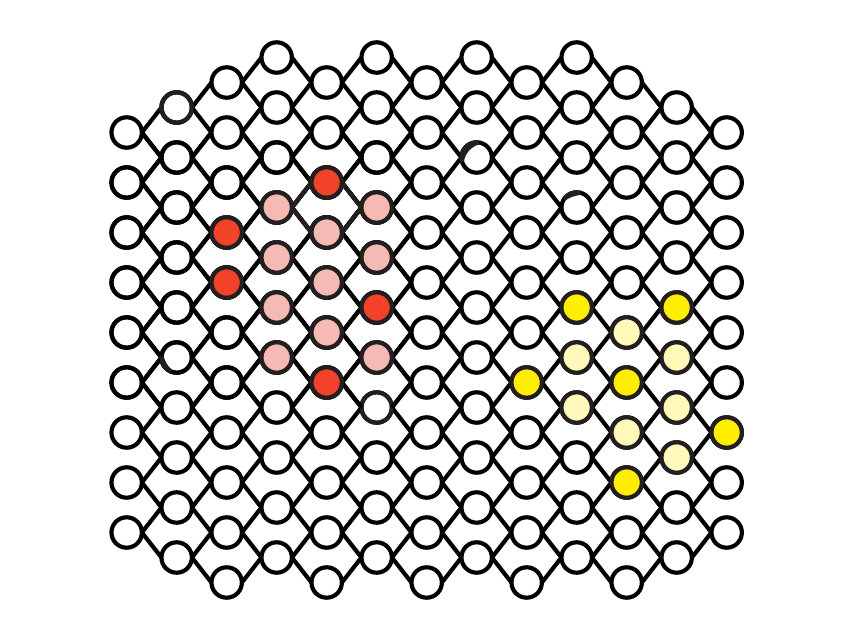}
	}
 	\hfill	

	\subfigure[]{
	\includegraphics[width=\linewidth]{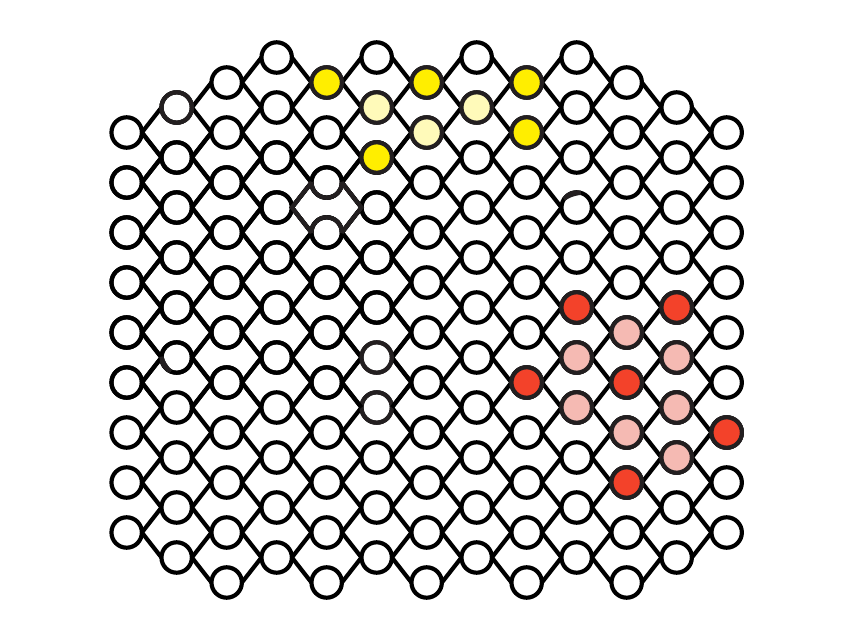}
	}
 	\hfill	
	\subfigure[]{
	\includegraphics[width=\linewidth]{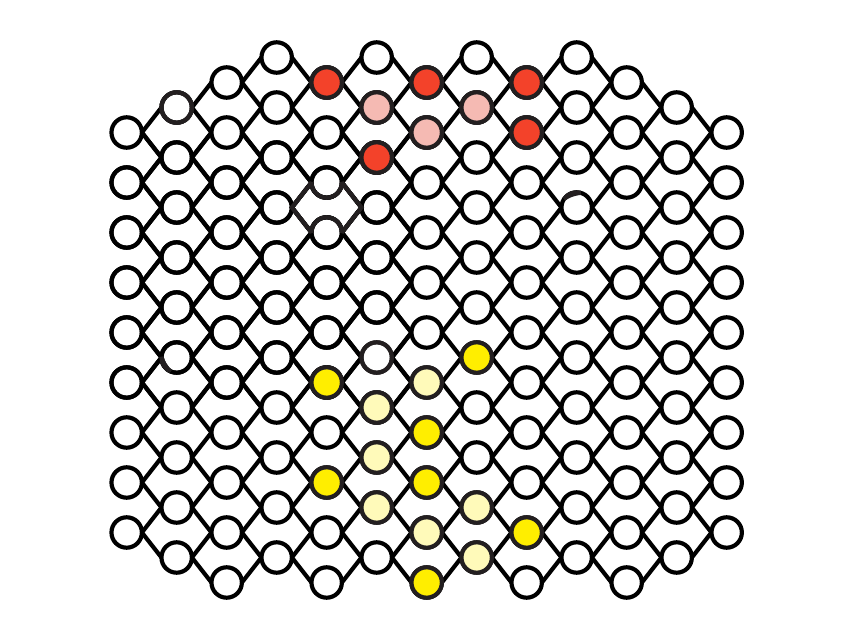}
	}	
\caption{Three cycles of episodic memory recall, showing retrieval cue (orange) and reactivation (yellow)}\label{fig:episodic-memory}
\end{figure}

A sequence of events can be coded as a set of cue-recall pairs. The ensemble of bursting neurons coding for a particular experience acts as memory retrieval cue to reactivate the next ensemble of bursting neurons. The bursting activity of these neurons would reactivate yet another ensemble of bursting neurons and the cycle continues, forming a stream of thoughts (figure~\ref{fig:episodic-memory}).

The reactivation sequence of neuron ensembles may not match with the coding sequence exactly. The neuron ensembles being reactivated by the hippocampus are constantly being merged with neurons being activated by current sensory stimuli. This allows current sensory stimuli and past experiences to be integrated, forming new episodic memories. Hence a different neuron ensemble may be reactivated from the same retrieval cue next time. The degree which current sensory information and past experiences can be integrated also depends on the underlying perceptual relations coded by cortical connections, as discussed in section~\ref{sec-percept-integration}.

The advantage of the episodic memory mechanism is that it allows learning to occur rapidly with little repetitive training. It allows any novel and significant sensory percepts and motor volitions to be bound, and reactivated when the same situation is subsequently encountered. The mechanism by which significance is coded by the brain is discussed further in section~\ref{sec-neuromodulation}.
\section{Neuromodulation}\label{sec-neuromodulation}

The classical formulation of reinforcement learning as a Markov decision process is to search for an optimal policy to associate states, actions, and rewards~\cite{Giryes2011}. It has been well understood that neuromodulation is the brain's mechanism to signal reward and punishment. The activation level of the various neuromodulators codes for different reward or punishment values. This  allows for learning of behaviours leading to rewards and avoiding punishments. Neuromodulation also gives rise to emotions, which includes varying degree of physiological responses in addition to the coding of reward/punishment values, such as the activation of the sympathetic nervous system to engage the "fight or flight" response.

There are many different types of neuromodulators. In particular, dopamine (DA), serotonin (5-HT), noradrenaline (NA), and acetylchloine (ACh) have been studied extensively for their roles in reinforcement learning and emotions. Neuromodulators are released at the synaptic cliff and act as neurotransmitters. Each neuromodulator has either excitatory amplification or inhibitory attenuation to synaptic strength and neuron activity, depending on the target. They are generated in different nuclei of the midbrain, and have pathways that extend across the whole cortex, thalamus, and hippocampus~\cite{Iwanczuk2015}.

Tomkin's affect theory outlines eight basic emotions modulated by different levels of DA, 5-HT and NA~\cite{Tomkins2008}. The theory suggests that both DA and 5-HT modulate different positive and negative emotions, and NA modulates the level of emotional arousal. While such theoretical model of emotion allows us to understanding the role of each neuromodulator, their exact influence from the prospective of reinforcement is more subtle and requires more detail discussion.

\subsection{Dopamine}\label{sec:dopamine}
DA has been understood to code for reward prediction error (PE)~\cite{Glimcher2011a}. Only the frontal cortex receives the DA signal, but not the parietal, temporal, and occipital cortices~\cite{Dahlstrom1964}. In the classical experiment of DA modulation, dopaminergic neurons were found to be active above the baseline when mice were given a reward, the unconditional stimulus (US)~\cite{Schultz1997}. The mice then learnt to associate that a conditional stimulus (CS), such as a tone, would precede the US. In subsequent trials, the CS incited DA activity while the US did not. Crucially, if no reward was given after the CS, DA activity dropped below baseline, coding for a negative reward PE.

In the hippocampus, the application of DA was found to convert post-before-pre long term depression (LTD) into long term potentiation (LTP) for spike timing dependent plasticity (STDP), even applied after a long time from the pairing~\cite{Brzosko2017}, meaning DA can strengthen the synaptic connection between two neurons irrespective of the precise spiking order.

\subsection{Serotonin}\label{sec:serotonin}
While the role of DA in reward PE is well understood, the role of 5-HT in reinforcement learning is less clear and more controversial. The 5-HT system receives inputs from the pre-frontal cortex (PFC), which is associated with reasoning and higher order executive functions, and the lateral habenula (LHb) which is associated with mediating aversive signals ~\cite{PollakDorocic2014, Zhou2017}. Studies have found that the level of 5-HT is correlated with expression of aversion following social defeat stress in mice~\cite{Challis2013}.

5-HT was initially hypothesised to inhibit behaviours leading to adverse outcomes~\cite{Soubrie1986}, and has been described as an opponent to DA~\cite{Daw2002}. However, studies have also found that 5-HT is linked to reward PE~\cite{Kranz2010}, complementary to the DA system. In fact, the DA and 5-HT system have projections targeting and exerting influence over each other~\cite{PollakDorocic2014}.

As such, 5-HT has been suggested to reflect mainly unsigned PE~\cite{Fischer2017}, or surprise. In studies with reversal learning tasks which a previously good stimulus is switched to a bad stimulus, the decline in DA activity was found to be faster and earlier, whereas the increase in 5-HT was slower and longer-lasting~\cite{Matias2017}. It was suggested that 5-HT thus plays a role in inhibiting reinforced behaviour to promote behavioural flexibility to allow for expectancy change.

\subsection{Noradrenaline}\label{sec:noradrenaline}
NA is one of the earliest neuromodulators to be studied extensively. Initially, NA was suggested to control arousal and wakefulness~\cite{Berridge2003}. Specifically, NA neurons were found to fire in tonic mode at a regular slow rate (\(\sim\)1Hz), and they fire in bursting mode in response to arousing~\cite{Devilbiss2000}, rewarding~\cite{Sara1991, Nieuwenhuis2005}, and aversive stimuli~\cite{Hirata1994, Bekar2008, Chen2007}. In other words, NA system responds to both reward and punishment, similarly to 5-HT. 

The NA system has also been associated with focus, attention, and performance~\cite{Berridge2003, Aston-Jones2005}. In the cortex as well as the cerebellum, NA was found to enhance the signal-to-noise ratio, by inhibiting spontaneous activity while no affecting the evoked response~\cite{Foote1975, Waterhouse1980, Woodward1991}. NA was also found to enhance the synchronisation of neuron activity, decreasing jitter in response latency and increasing synchronous precision in the sub-millisecond range~\cite{Lecas2004}. The role of NA in selective attention was further supported by experiments in rats with blockage of receptors to NA in the olfactory bulb showing impaired ability to discriminate between closely matching odours~\cite{Doucette2007}.

In addition to the above transient influences, NA plays a critical role in memory formation and neural plasticity. In the hippocampus, burst firing of NA neurons has been shown to facilitate LTP~\cite{ODell2010a}. At the cellular level, NA was found to be linked to the transition from short term potentiation that lasts for a few hours, to long term potentiation that requires synthesis of new protein~\cite{Straube2003, Gelinas2005}. Rats injected with NA receptor antagonist 2 hours after learning showed amnesia, whereas there was no effect if the injection was applied immediately after learning, suggesting the crucial role of NA in turning short-term memory to long-term memory \cite{Sara1999, Tronel2004}.

In addition to memory formation and consolidation, NA also plays a role in memory retrieval. Injection of NA receptor antagonist in rat was shown to impair memory retrieval 24 hours after training, but interestingly not 1 hour or 1 week after training~\cite{Murchison2004}. During memory retrieval, the NA system is also reactivated by the amygdala, the brain region responsible for emotional memory, resulting in similar physiological responses as during the coding phase~\cite{Sterpenich2006}.

\subsection{Acetylchloine}\label{sec:acetylchloine}
ACh's role seems to be complementary to the NA system, and it has also been linked to focus, selective attention, and wakefulness. ACh neurons project to all cortical regions, as well as the the hippocampus~\cite{Woolf1991}. ACh activity in the magnocellular preoptic nucleus and Substantia innominata is higher with cortical gamma activity during wakefulness and rapid eye movement sleep, and it has been suggested that ACh is activated by the NA system~\cite{Deurveilher2011}. In rats, a spike in ACh level in the medial prefrontal cortex was observed when the rat oriented its attention to a sensory cue~\cite{Parikh2007a}. The concentration of ACh in prefrontal cortex more than doubled when the rat was performing tasks requiring sustained attention in another experiment~\cite{Sarter2006}. In a similar experiment, novel stimulus was found to induce significant frontal cortical and hippocampal ACh release, whereas another group previously trained with the same stimulus did not show the same increase~\cite{Acquas1996}. Interestingly, a third group also trained with the same stimulus but as a CS to a foot shock also result in ACh increase, without observing the same habituation effect. This shows that ACh correlates to both novelty and fear conditioning.

The ACh system has bidirectional projection from and to the pre-frontal cortex~\cite{Gaykema1991}, as well as receiving inputs from the DA system~\cite{Gaykema1996}. The frontal cortex is the brain region carrying higher order executive function and reasoning~\cite{Fuster2000}. In particular, the anterior cingulate cortex (ACC) of the frontal cortex was found to be linked with error processing, conflict resolution, and losses of reward~\cite{Botvinick2004}. It was found that the loss of ACh inputs to the medial pre-frontal cortex neurons in rats attenuated their firing rates in a visual distraction test, but this did not affect sustained attentional performance~\cite{Gill2000}, suggesting a role ACh plays in error detection.  

In the hippocampus, ACh was found to cause LTD in active synapses, regardless of the spiking order~\cite{Brzosko2017a}. It was suggested that ACh induced hippocampal LTD allows negative outcomes to be forgotten, allowing for flexible learning~\cite{Zannone2017}. ACh was found to be higher in the hippocampus during active waking when the animal was actively exploring the environment, and lower during slow wave sleep and quiet waking when the animal was less active. A general correlation between the amplitude of hippocampal theta oscillations and ACh level was observed~\cite{Monmaur1997}. Thus ACh is suggested to have an effect of suppressing hippocampus memory recall so that the coding of current sensory information is not disrupted by previously stored information~\cite{Hasselmo1999}.

\doublerulesep 0.1pt
\begin{table*}[t]
  \centering
 \linespread{1.7}{ {\footnotesize
  \caption{Reinforcement learning rules under different scenarios}\label{tab:general-reinforcement}
\vspace{1em}
  \begin{tabular}{ccc}
  \hline
\noalign{\smallskip}
    & \textbf{Pleasure/reward} & \textbf{Pain/punishment} \\
\noalign{\smallskip}
  \hline
    \textbf{Positive prediction error} & Learning state/action to REINFORCE reward & Learning state/action to AVOID punishment \\
    \multirow{2}{*}{\textbf{Negative prediction error}} & Unlearning prior state/action to reward & Unlearning prior state/action to punishment \\
     & Learning new state/action to AVOID non-reward & Learning state/action to REINFORCE non-punishment \\
  \hline
\noalign{}
  \end{tabular}
  }}
\end{table*}

\subsection{Emotional memory and amygdala}
Section~\ref{sec:dopamine} to section~\ref{sec:acetylchloine} reviewed the influences and mechanisms of the four main neuromodulation systems. However, what is the neural mechanism that triggers the activity of the different neuromodulation systems and associated emotions? The most basic form of emotion arises from homoeostatic control, such as the maintenance of body temperature, blood sugar level, etc~\cite{Denton2012}. To arouse homoeostatic emotion, activation of the corresponding sensory neurons is required, such as the various nociceptors for the perception of pain~\cite{Criado2010}, or receptors in gastrointestinal tract for the sensation of hunger~\cite{Marieb2014}.

In addition to homoeostatic emotions, any sensory percepts can be bound to an emotional state through Pavlovian classical conditioning~\cite{Pavlov1927}, and such process is modulated by the amygdala. Damage to the amygdala was found to impair Pavlovian fear conditioning~\cite{Ressler2003a}. The amygdala reactivates the corresponding neuromodulator in the midbrain as when it was coded for the CS. Such process happens rapidly and is not dependent on awareness or attentional focus~\cite{Whalen1998, Anderson2013a}. Amygdala activity correlates with the level of emotional arousal, and this affects how well the emotional memory is retained~\cite{Pare2002}.

The emotional memory function of the amygdala is often linked to the episodic memory function of the hippocampus. Episodic memory is often richer with finer details when under a state of high emotional arousal, a phenomenal known as flashbulb memory~\cite{Brown1977}. Some suggest the amygdala simply modulates the memory consolidation of other brain areas, such as the hippocampus through the NA system~\cite{Cahill1999}, while some suggest that it is the site of emotional storage itself~\cite{Fanselow1999}. Regardless, it was found that patients with amygdala damage failed to show a normal fear response in a fear conditioning experiment but they are able to predict the US from the CS, whereas patients with hippocampus damage failed to make the prediction but showed fear response. Finally, patients with damage to both the amygdala and hippocampus failed at both, suggesting a differential role for amygdala and hippocampus for emotional and episodic memory~\cite{Bechara1995}.

\subsection{Selective attention and action selection}
Section~\ref{sec-percept-integration} discussed that synchrony ensembles compete for dominance of awareness, and it was suggested that the relative size and firing rate of the ensembles are the key factors to determine which ensemble becomes dominant. This form of selective attention is often described as bottom-up selective attention, which is entirely driven by sensory saliency~\cite{Posner1990}.

In addition, top-down selective attention is when attention is driven by awareness and goal volitions. It is understood that neuromodulation plays an important role in top-down selective attention. In particular, the effect of enhancing the signal-to-noise ration of cortical neurons by the NA system is one principle mechanism of top-down selective attention modulated by emotional arousal~\cite{Foote1975, Waterhouse1980, Woodward1991}. In addition, the ACh system is also suggested to be involved in top-down selective attention~\cite{Sarter2006}. 

Action selection is closely linked to, but not identical to selective attention. Attention can be paid to a particular set of sensory stimuli, but the decision of which action to take, and whether or not to take an action, depends on the executive function of the frontal cortex. One can focus attention towards a stimulus while not taking any actions. The basal ganglia is another brain region linked to action selection. The basal ganglia receives inputs from different cortical regions in a topological manner, with inputs from different cortical areas targeting different basal ganglia nuclei. It also sends largely inhibitory output back to the cortex topologically via the thalamus, as well as the cerebellum that is responsible for controlling smooth motor movements, forming loops of movement control between the cortex and the cerebellum.~\cite{Middleton1994, Middleton2000, Middleton2000a}. The basal ganglia also receives inputs from the ACh, 5-HT and the DA systems, indicating the role for neuromodulation in action selection~\cite{DiMatteo2008}. 

\subsection{General reinforcement learning rules}
Given the many different and seemingly conflicting roles of each neuromodulation system reviewed above, it may be difficult to draw a definitive conclusion on the exact mechanism of neuromodulation to allow for reinforcement learning in the brain. Many attempts have been made to develop computational models for reinforcement learning based on understanding of neuromodulation in the brain, most notably the temporal difference learning model based on the finding of the DA coding of reward PE~\cite{Glimcher2011}. Other models have attempted to incorporate other neuromodulators, suggesting 5-HT controls the time scale of reward prediction, NA controls the randomness in action selection, and ACh in control of the learning rate~\cite{Doya2002}.

based on the evidence reviewed above, there may be more complex sub-processes which govern the learning and selection of reinforced or aversive behaviours, under positive and negative prediction error of both reward and punishment. A proposal of the learning rules under different scenarios is summarised in table~\ref{tab:general-reinforcement}.

Under the current proposal, only novel stimuli result in awareness by neuron burst firing. This guarantees all stimuli that result in awareness are either from an unexpected event, a positive prediction error, or an expected event not being observed, a negative prediction error. Learning of the associated states or actions to the rewards or punishments should happen under both scenarios. However, for a negative prediction error, there must also be a mechanism for the brain to unlearn prior associations, otherwise learning becomes saturated very quickly, and no new learning can occur. In addition, it is also important to learn the association between states and actions leading to punishments, in order to avoid them in subsequent trials.

It is a basic assumption that the brain needs to learn to reinforce actions leading to reward and avoiding punishment. However, the very idea of reward and punishment is in fact subjective. From a sensory perspective, there is only pleasure and pain. To obtain the higher level percept of rewards and punishments, prior learning of pleasure/pain association by the amygdala, and reasoning by the frontal-cortex is needed.

\doublerulesep 0.1pt
\begin{table}[htb]
  \centering
 \linespread{1.7}{ {\footnotesize
  \caption{Differential roles of neuromodulators}\label{tab:neuromodulator-roles}
\vspace{1em}
  \begin{tabular}{ccc}
  \hline
\noalign{\smallskip}
    & \textbf{Learning/reinforcement} & \textbf{Unlearning/aversion} \\
\noalign{\smallskip}
  \hline
    \textbf{Memory} & NA & ACh \\
    \textbf{Action selection} & DA & 5-HT \\
  \hline
\noalign{}
  \end{tabular}
  }}
\end{table}

From this perspective, the seemingly conflicting role of each neuromodulator may be explained by their differential roles in learning/unlearning, which concerns memory and plasticity, as well as reinforcement/aversion, which concerns action selection. Table~\ref{tab:neuromodulator-roles} summarises the role of each neuromodulator under this hypothesis. The role of NA in episodic and emotional memory function has many pieces of psychological and neurophysiological evidence~\cite{ODell2010a, Straube2003, Gelinas2005, Sara1999, Tronel2004}. The role of ACh in unlearning has been suggested based on evidence of induced hippocampal LTD by ACh~\cite{Brzosko2017a, Zannone2017}, as well as the link between the ACh system and ACC, the error detection region of the brain~\cite{Botvinick2004, Gill2000}. The role of DA and 5-HT in reinforcement/aversion, as opposed to simple reward/punishment is more controversial, but has also been suggested before~\cite{Cools2011}. Regardless of the exact role of each neuromodulator in the brain, it is still worth investigating the advantage in reinforcement learning using the rules in table~\ref{tab:general-reinforcement} and the signals in table~\ref{tab:neuromodulator-roles}. It may well be the case that the brain uses a combination of two or more neuromodulators to create each of the signals outlined in table~\ref{tab:neuromodulator-roles}.

\section{Learning and plasticity}\label{sec-learning}

The percept integration and episodic memory mechanism discussed in section~\ref{sec-percept-integration} and~\ref{sec-episodic-memory} allows for rapid learning of novel and significant associations. While such a learning mechanism is fast, it requires awareness and is slow at recall. In all previous sections, the underlying cortical connections are assumed to have already represented a level of perceptual relations of the underlying real world correlation and causality. However, how do the cortical connections develop such perceptual relations in the first place?

As the brain develops from infancy to adulthood, cortical connections are constantly adapting as one interacts with the environment. One leading theory of neuron plasticity is the spike timing depended plasticity (STDP), which is observed in the cortex as well as the hippocampus~\cite{Song2000}. STDP allows for the strengthening of synaptic connections if a pre-synaptic neuron fires just before post-synaptic neuron, resulting in long term potentiating (LTP). On the other hand, the synaptic strength is decreased if the firing sequence is reversed, resulting in long term depressing (LTD).

The binding by synchrony mechanism described in section~\ref{sec-percept-integration} is crucial for the correct direction of synaptic weight update to occur. Without frequency synchronisation, phase shift between two neurons occurs, leading to both LTP and LTD and no overall synaptic weight update.

The theory of systems consolidation of memory suggests that memory is first stored as episodic memory in the hippocampus, and is then slowly transferred to the cortex as semantic memory and procedural memory~\cite{Squire1995, Frankland2005}. Rapid eye movement (REM) sleep is suggested to enhance this process by replaying the activation pattern of the corresponding cortical regions coded in episodic memory by the hippocampus, with dreaming being a by-product of such a process~\cite{Walker2005, Stickgold2005, Wamsley2010}. However, evidence also suggests that with enough repetition, patients without the hippocampus can still acquire certain level of semantic and procedural memory~\cite{Squire1986}.

\begin{figure}[htbp]
 	\centering
 	\subfigure[]{
	\includegraphics[width=\linewidth]{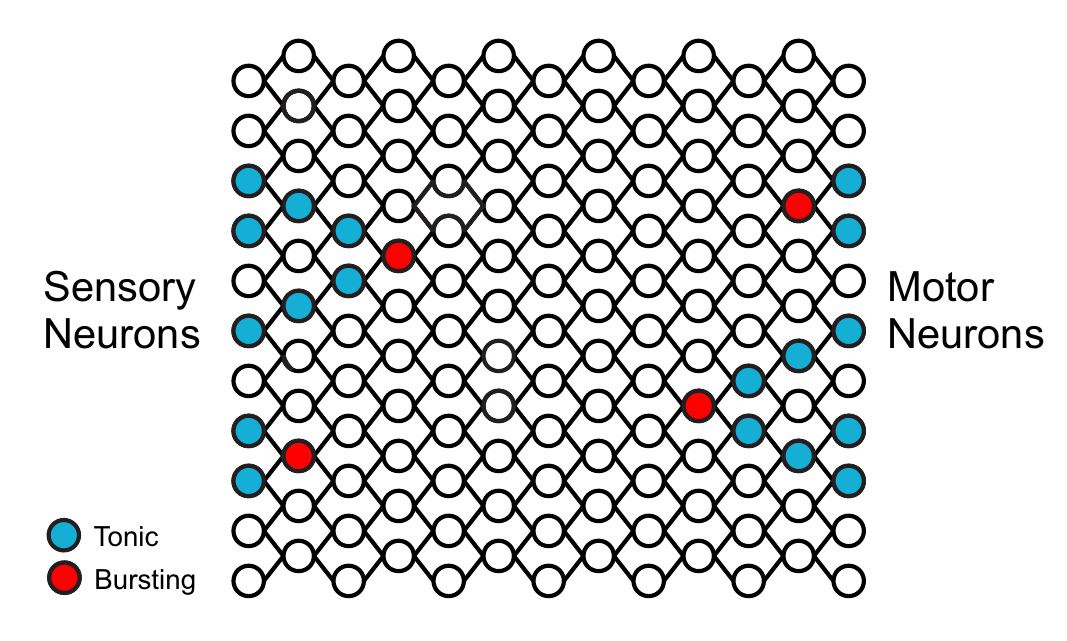}
	}
 	\hfill
	\subfigure[]{
	\includegraphics[width=\linewidth]{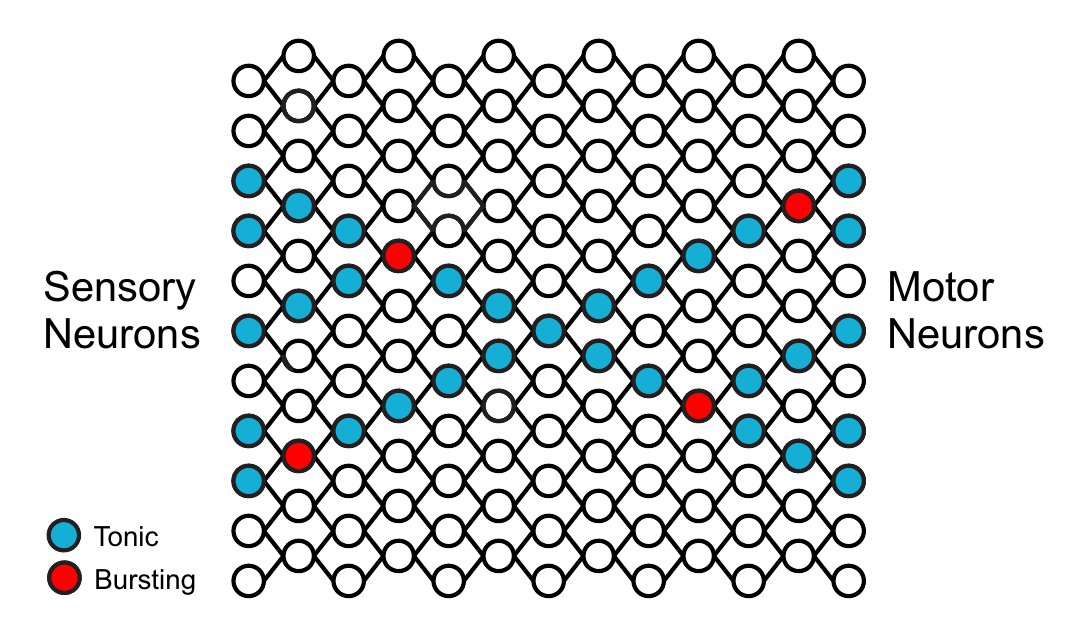}
	}
 	\hfill	
	\subfigure[]{
	\includegraphics[width=\linewidth]{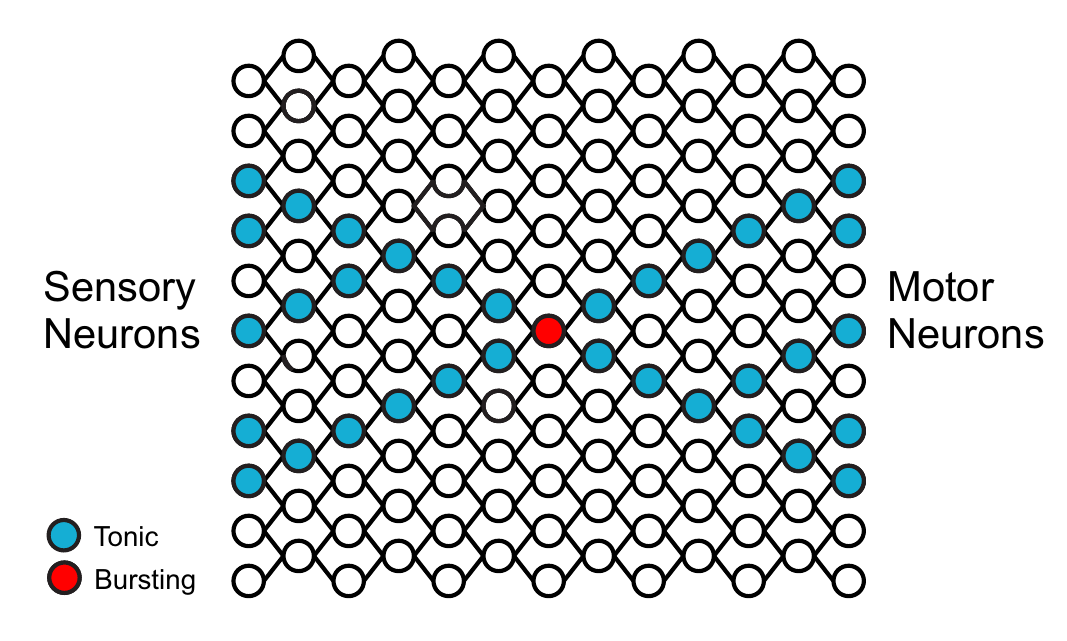}
	}	
\caption{Learning mechanism of cortical neurons. a) Several sensory percepts and motor volitions result in awareness. b) Formation of synchrony ensemble and coded by episodic memory. c) Through strengthening of cortical connections by STDP, the number of bursting neurons decreases, resulting in a higher level of perceptual abstraction with less awareness.}\label{fig:learning}
\end{figure}

With repetitive training and the mechanism of memory consolidation by the hippocampus, cortical connections coding for percepts and volitions are strengthened by STDP. It is proposed that through strengthening of cortical connections in a synchrony ensemble stored in episodic memory, new neurons become bursting while previous bursting neurons become tonic. The number of bursting neurons thus decreases as cortical connections are strengthened, and the bursting neurons move up a higher level of perceptual abstraction (figure~\ref{fig:learning}).

With such a cortical learning mechanism, awareness becomes less available for familiar circumstances and actions. However, it allows the brain to build associations between ever higher levels of abstraction. This allows perceptual model of arbitrary depth to be learnt by the brain, by building on top of the highest level of current perceptual abstraction incrementally.

\section{Reasoning}\label{sec-reasoning}
The exact neuromechanism of reasoning in the brain is still largely elusive. Reasoning as a cognitive process differs from learning as it involves the goal directed integration of prior knowledge and current information, and it requires recurrent and feedback interactions. Chapter~\ref{sec-novelty-representation} described binding by synchrony as a neuron mechanism to integrate prior knowledge and current sensory information. However, it lacks a clear explanation of how logical reasoning can be formed by the synchronised interactions of neuron ensembles. There are many formulations and theories for logical reasoning, from deductive/inductive reasoning to Modus Ponens/Modus Tollens. However, experimental data indicates that the behaviour of human reasoning does not follow strict theoretical reasoning frameworks, and no particular framework or theory can wholly and adequately explain human reasoning~\cite{Khemlani2012}. In other words, human reasoning is seemingly not entirely logical, yet it is powerful in that it allows human to make decisions under complex situations with often incomplete information. The seemingly stochastic nature of human reasoning inspires many Bayesian methods to understand the brain~\cite{Doya2007}. The Bayes theory is a useful statistical theory to understand many probabilistic systems in nature, however it does not explain the underlying neuromechanism which causes the probabilistic property.

In computer science, machines follow strict logical rules. Most modern computers use the Von Neumann architecture that separates logic operations, memory, and control. All logic operations can be accomplished by the combination of three basic logic gates, AND, OR and NOT. Pre-defined logic operations built by these logic gates allow general computation to be performed by a control unit executing program instructions stored in memory. Equivalent logic gates can also be found in the brain, with excitatory neurons firing depended on the inputs from one AND/OR many upstream neurons. The inhibitory interneurons on the other hand acts as the equivalent of NOT gates in the brain. This is however where the similarity ends, as the architecture and reasoning process of the brain is vastly different from that of the computer. Memory and logic operations are intertwined in the brain, and the logic operation performed by each local neuron circuitry entirely depends on its inputs, outputs, and the synaptic connections between neurons. It can be said that the organisation of the logic units in the brain is more analogical to a field-programmable gate array (FPGA), which allows logic gates to be reorganised to implement new logic operations. The brain employs a different neuron circuitry in different part of the cortex depending on the task at hand, and constantly reorganises its neuron connections to perform new logic operations to better suit the tasks.

As described in section~\ref{sec-learning}, the brain reorganises its neuron connections by the STDP update rule modulated by neuromodulation. STDP allows associative learning to occur, thus allowing the AND and OR operations to be updated. This forms of logic operation allows generalisation and ever higher levels of abstraction in the perceptual map. However, the forming of the NOT operation needs to be guided by an aversion signal, as described in section~\ref{sec-neuromodulation}.

Recent work in deep meta reinforcement learning (RL) suggests that the activation pattern of the pre-frontal cortex (PFC) neurons allows for a fast model based RL procedure when the meta RL is trained model free~\cite{Wang2018}. In other words, the underlying cortical connections establish the appropriate perceptual models of sensory stimuli, and the activation pattern of PFC neurons allows fast logical reasoning to occur based on these models.
\section{Evidence for current proposal}\label{sec-evidence}

The current proposal relies on the hypothesis that the neuron bursting mode codes for novelty and is what results in awareness. There is currently no consensus within the neuroscience community regarding the relevance and functions of neuron bursting mode in contrast to tonic mode~\cite{Zeldenrust2018}. Many neuron types exhibit bursting behaviours in brain regions involved in the current proposal, including the cortex~\cite{Baranyi1993, Gray1996, Gray1997}, hippocampus~\cite{Traub1981, Miles1986}, thalamus~\cite{Jahnsen1984, Williams1997}, and neuromodulation systems~\cite{Wang1981, Hyland2002, Devilbiss2000}. In particular DA neuron bursting is found to be associated with a larger dopamine release in their target as well as with reward related stimulus~\cite{Cooper2002a}.

The functional role of neuron bursting was investigated in the weakly electric fish~\cite{Gabbiani1996, Krahe2004}, and it was found that bursts are used in every level of sensory processing of the electric fields. It was suggested that bursts play a feature detection role, and are more reliable than single spikes~\cite{Metzen2016}.

However, question has been raised if bursts allow a more reliable form of information transmission than single spikes in tonic mode, then what is the functional role of single spikes? It was suggested that single spikes in tonic mode are only relevant in conjunction with spikes from other neurons~\cite{Zeldenrust2018}.

Multiplexing code was also suggested for the functional role of bursts. It was found that thalamocortical relay neurons use such a code, where information about the stimulus is conveyed in the burst size, in the burst onset time and in spike timing within bursts. In addition, thalamic neurons bursting is suggested to act as a wake-up call to activate their cortical target~\cite{MurraySherman2001}.

Pyramidal neurons in hippocampus have been called place cells, as experiments show that these neurons only fire when an animal is at specific locations~\cite{OKeefe1971}, called place-fields. It was found that place-fields are more accurately defined by bursting spikes only, as suppose to both tonic and bursting spikes~\cite{Otto1991}. Evidence suggests that the hippocampus relies only on bursts for internal information transfer~\cite{Buzsaki2012, Xu2012}, supporting the current proposal of bursting neurons as feature detectors of high level abstraction.

Bursts have also been found to modulate plasticity and learning in the hippocampus as well as the cerebellum. It was found that strong theta wave stimulation excitatory postsynaptic potentials (EPSP) evoked both bursts and LTP in hippocampal pyramidal neurons, whereas weaker stimulation evoked single spikes but did not induce LTP~\cite{Thomas1998}.
\section{In relation to back propagation}\label{sec-back-propagation}

Similar to biological neural networks, deep neural networks have multiple layers of non-linear processing units that code for features in a hierarchical manner~\cite{Lecun2015a}. In particular the convolution neural network was inspired by the information processing hierarchy of the biological visual cortex~\cite{Fukushima1980}. However, successful learning of deep neural networks relies on the error back-propagation algorithm and supervised learning~\cite{Rumelhart1986, LeCun1998}. Many researchers have suggested that the error back-propagation algorithm is not biologically feasible, and more often the brain performs unsupervised learning~\cite{Crick1989}.

The learning efficiency of the back-propagation algorithm, as defined by the number of training samples required, is low compared to the learning efficiency of biological learning systems. Each additional layer of non-linearity in a deep neural network requires yet more training samples to prevent overfitting. In contrast, humans can learn with just a few training samples~\cite{Lake2016, Cox2014}. In addition, state of the art deep neural networks no longer employ sparse coding~\cite{HonglakLeeAlexisBattleRajatRaina2006}. This is equivalent to all neurons being active and contribute to the outcome proportionally. In contrast, coding in the brain is sparse~\cite{Foldiak2002} as information is coded as spikes, allowing for much higher energy efficiency.

Despite the clear advantage of biological learning rules, attempts to use STDP on artificial spiking neural networks experienced difficulty for networks beyond a few layers. They tend to perform worse than networks trained with back-propagation algorithm with the exact same architecture~\cite{McClelland2006}.

It has been suggested that pure STDP does not address the credit assignment problem, which is the problem of determining which neuron's activity has the highest contribution to the desired outcome, in order to strengthen the corresponding connections appropriately~\cite{Minsky1961}. Back-propagation algorithm addresses this problem by back-propagating the error gradient from the output side towards the input side, and updates the weights in each layer accordingly.

The current proposal addresses the credit assignment problem as follows:

1. First it allows the highest level of abstraction of sensory stimuli and motor commands to be identified by neurons in bursting mode.

2. The link between the most abstract representation of sensory percept (state), motor volition (action), and desired outcome (value) is identified by the binding by synchrony, episodic memory and the neuromodulation mechanism of the brain.

3. Through strengthening of cortical connections by repetition and memory consolidation mechanism, the bursting neurons move to an even higher level of perceptual abstraction.


The current state of the art artificial neural networks can have as many as ten thousand layers of neurons with 128 neurons per layer~\cite{Xiao2018}. In a human, is estimated that there are around 16 billion neurons in the cortex, and 86 billion neurons in total~\cite{Herculano-Houzel2009}. As mentioned, not all neurons in the brain are active at all time. In addition, the number of neurons information passes through from sensory input side to motor output side can vary depending on the task, as the thalamus act as the relay circuit to the entire cortex. Using the back-propagation algorithm to train artificial neural networks with billions of neurons would be challenging due to the vanishing gradient problem, which the error gradient becomes too small as it is propagated back through more layers~\cite{Bengio1993}. The current proposal allows synaptic weight updates to be carried out only between a much reduced number of neurons coding for high-level percepts, instead of updating all neuron connections simultaneously as is done with the back-propagation algorithm. This is how the current proposal is suggested to allow learning of perceptual abstraction of arbitrary depth, by updating only a few layers at a time at the most abstract levels.


\section{Conclusion}\label{sec-conclusion}

In this paper a theoretical model is proposed for the biological learning mechanism as a general learning system. The key hypothesis is that the underlying information processing mechanism which results in awareness is what allows biological neural networks to exhibit general intelligence. The core of the theory draws from the insight that there are two separate mechanisms for information processing in the brain, with one resulting in awareness and one being subliminal. While both modes of information processing in the brain can perform tasks with a similar level of complexity, the key difference is that learning of new tasks often requires awareness before subliminal processes can take over.

It is further proposed that the brain has the evolutionary purpose of processing novel information. Novelty is viewed as a combination of familiar information which has not previously been experienced in a particular configuration. Therefore the evolutionary function of the brain is to determine the appropriate response for the novel combination of familiar information that maximises reward and minimises punishment. 

The bursting and tonic mode of many neuron types are further proposed to be these two modes of information processing respectively. Under this hypothesis, bursting neurons represent the highest level of sensory and motor abstraction, and the ensemble of bursting neurons together represent the novelty being observed.

Binding by synchrony, episodic memory, neuromodulation, STDP, and the memory consolidation mechanism all have existing neurological and psychological evidence, An attempt has been made to unify all these key cognitive mechanisms with the theory of information processing by bursting neurons. Many fine details still need further proposals and investigation.

\bibliographystyle{unsrt}
\bibliography{ref}
\end{document}